\documentclass[sigconf]{acmart}
\usepackage{xcolor}
\usepackage{multirow}
\AtBeginDocument{%
  \providecommand\BibTeX{{%
    \normalfont B\kern-0.5em{\scshape i\kern-0.25em b}\kern-0.8em\TeX}}}

\copyrightyear{2023}
\acmYear{2023}
\setcopyright{rightsretained}
\acmConference[FIRE 2023]{Forum for Information Retrieval Evaluation}{December 15--18, 2023}{Panjim, India}
\acmBooktitle{Forum for Information Retrieval Evaluation (FIRE 2023), December 15--18, 2023, Panjim, India}\acmDOI{10.1145/3632754.3632759}
\acmISBN{979-8-4007-1632-4/23/12}

\begin{document}

\title{Are we describing the same sound? An analysis of word embedding spaces of expressive piano performance}
\author{Silvan David Peter}
\email{silvan.peter@jku.at}
\orcid{0009-0000-8328-291X}
\affiliation{
  \institution{Johannes Kepler University}
  \city{Linz}
  \country{Austria}
}

\author{Shreyan Chowdhury}
\email{shreyan.chowdhury@jku.at}
\orcid{0000-0003-1171-4036}
\affiliation{
  \institution{Johannes Kepler University}
  \city{Linz}
  \country{Austria}
}

\author{Carlos Eduardo Cancino-Chac\'on}
\email{carlos_eduardo.cancino_chacon@jku.at}
\orcid{0000-0001-5770-7005}
\affiliation{
  \institution{Johannes Kepler University}
  \city{Linz}
  \country{Austria}
}

\author{Gerhard Widmer}
\email{gerhard.widmer@jku.at}
\orcid{0000-0003-3531-1282}
\affiliation{
  \institution{Johannes Kepler University}
  \city{Linz}
  \country{Austria}
}

\renewcommand{\shortauthors}{Peter et al.}

\begin{abstract}

Semantic embeddings play a crucial role in natural language-based information retrieval. 
Embedding models represent words and contexts as vectors whose spatial configuration is derived from the distribution of words in large text corpora. 
While such representations are generally very powerful, they might fail to account for fine-grained domain-specific nuances.
In this article, we investigate this uncertainty for the domain of characterizations of expressive piano performance.
Using a music research dataset of free text performance characterizations and a follow-up study sorting the annotations into clusters, we derive a ground truth for a domain-specific semantic similarity structure.
We test five embedding models and their similarity structure for correspondence with the ground truth.
We further assess the effects of contextualizing prompts, hubness reduction, cross-modal similarity, and k-means clustering.
The quality of embedding models shows great variability with respect to this task; more general models perform better than domain-adapted ones and the best model configurations reach human-level agreement.

\end{abstract}






\maketitle

\section{Introduction}

Semantic embeddings are a foundational concept in Natural Language Processing (NLP).
NLP embedding models map words and their contexts to high-dimensional vector spaces while encoding as much of the semantic information as possible.
Computers process such numerical data more readily than text, and vector spaces enable simple yet powerful similarity representations.
Semantic embeddings enable a wide a variety of downstream tasks, from retrieval to classification to context-enhanced few shot learning.
State-of-the-art (SOTA) embedding models are trained on vast datasets of natural language covering many domains and disciplines, underpinned by a general distributional hypothesis --- that words with similar meanings occur in similar contexts. 

This hypothesis and its corresponding inductive bias lead to one of the major open challenges related to semantic embeddings:
whether context dependence, polysemy, and fine-grained domain-specific idiosyncrasies are adequately represented by general purpose semantic embeddings.
Specialized domains of language use such as talk of specific arts 
might exhibit incommensurable associations and similarites, e.g., a bright piano sound evokes different meanings than a bright student.
While specific domains and their possibly nuanced differentiations do occur in large datasets, they only occur in their specific context, where the distributional hypothesis runs counter to different encoding.
In other words, if certain emotionally dissimilar adjectives (happy, sad) only occur in similar contexts, they end up close in the embedding space despite opposing meanings.

\begin{figure*}[t]
\begin{center}
%  trim={<left> <lower> <right> <upper>}
\includegraphics[trim={0 100pt 0 80pt},clip,width=0.95\linewidth]{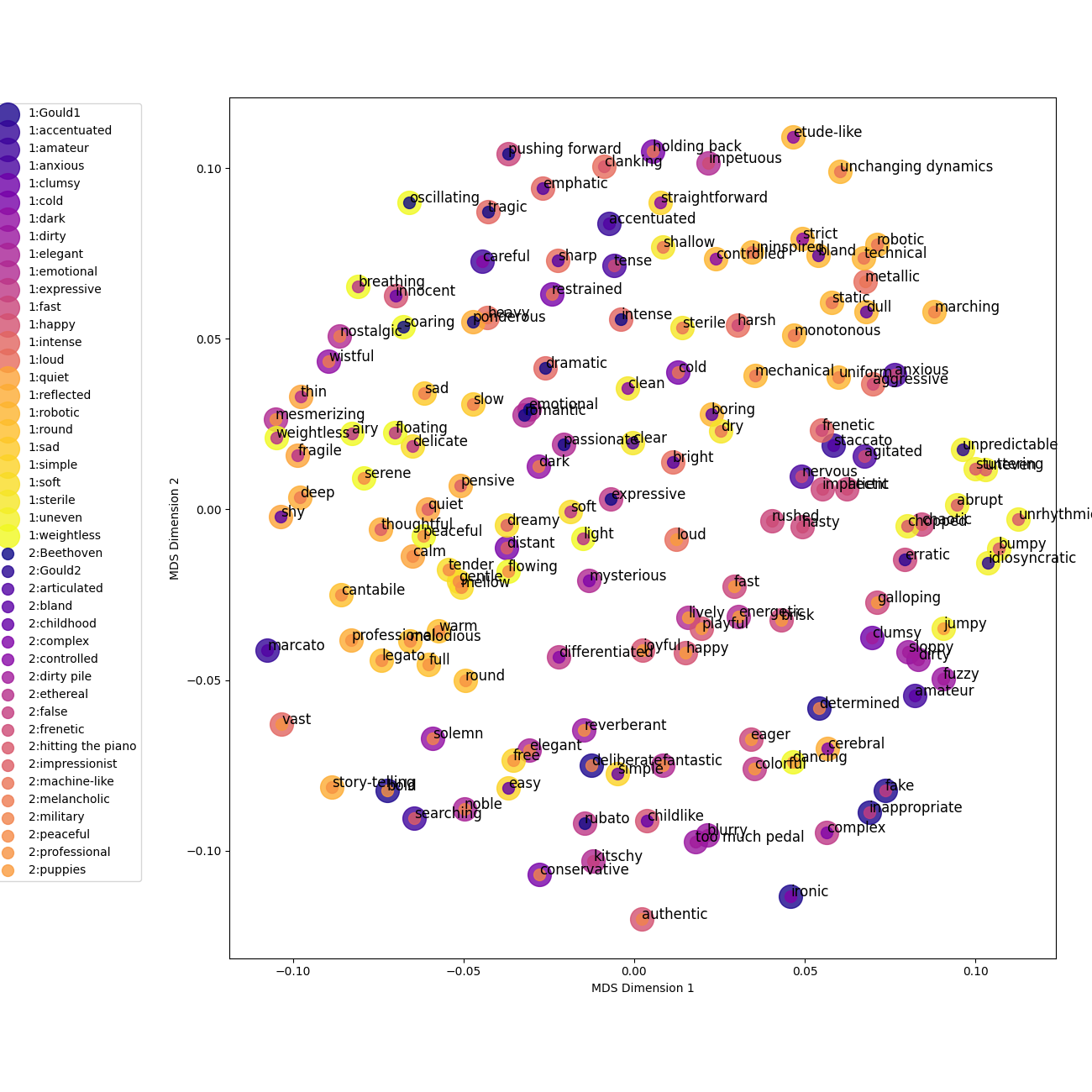}
\caption{Multidimensional Scaling (MDS) of the term data based the equally weighted pile group one,  pile group two, and performance similarities. 
The legend on the left lists all piles of both groups, first the group number, then the names the musicians assigned them. 
Each term of the 150 in our ground truth data is shown in the scatter plot to the right and colored by the two piles it was sorted into, one for group one (large dots), one for group two (small dots).
The musicians did not rate any similarities between piles, the color progressions for the piles do not encode closeness.
}  
\label{fig:mds_main}
\end{center}
\end{figure*}

\begin{figure*}[t]
\begin{center}
\includegraphics[trim={0 0 0 0},clip,width=0.95\linewidth]{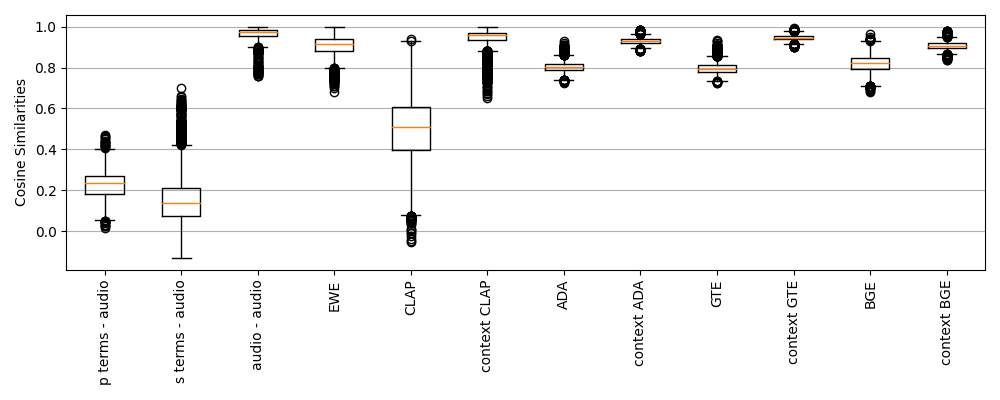}
\caption{Box plots of distributions of pairwise similarities in various embedding spaces. 
The main embedding models tested are labelled as EWE, CLAP, ADA, GTE, and BGE~\cite{li2023general,OpenAI_2022,BAAI_2022,elizalde2023clap,agrawal2018learning}.
The three leftmost distributions relate to cross-modal audio and text embeddings as discussed in Section~\ref{sec:audio} and the distributions labeled with "context" are addressed in Section~\ref{sec:context}.
}  
\label{fig:simdist}
\end{center}
\end{figure*}

In this article, we address the question whether general semantic embeddings can recover similarity relations in a specific domain of language use.
In particular, we are interested in adjectival spaces used in the characterization of expressive performance of Western classical solo piano music.
The underlying motivation is the possibility of expressivity- or emotion-based music retrieval using intuitive verbal queries, which would be a valuable and sought-after service in the digital music world.

Characterizations of expressive performance get at the finest details of performance technique, expression, timbre, emotions, metaphors, and associations.
Note that it's crucially not the piece that is being described by such characterizations, but its expressive interpretation and rendition.

Characterizations of expressive performance are both highly specific as well as very important to domain experts such as performers, teachers, and committed listeners.
In fact, a crucial skill for aspiring performers consists in developing a sensibility as well as a language for performance nuances.
Likewise, discriminating classical music lovers are highly sensitive to interpretation differences and can be very articulate in describing aspects of a performance that they don't like.

We use a dataset of terms used for the characterization of expressive performance in a large scale listening study.
The dataset is annotated with similarity clusters of 150 terms 
created by two groups of domain experts in Western classical music performance.
This data gives us an ecologically valid adjectival space of domain specific terms along with expert-annotated similarity annotations --- an ideal experimental reference for general embedding spaces.

In this article, we take this reference similarity space of 150 terms related to the domain of expressive performance characterization, and compare it with embeddings for these terms derived from five embedding models by means of precision at k metrics.
Furthermore, we present experiments investigating several factors affecting the embedding spaces: adding context to terms, reducing hubness in the embedding spaces, comparing audio to text embeddings, and a comparison of different clusterings in the embedding spaces.

\section{Related Work}\label{sec:related}

The literature on semantic embeddings is vast and multifaceted, with many tasks and benchmarks making use of suitable language representations.
For a recent overview and online benchmarking results, we refer to the Massive Text Embedding Benchmark (MTEB)~\cite{muennighoff2022mteb,MTEBLeaderboard}.

In our work, we investigate five models which we introduce in the following.
These models cannot represent the whole of the state of the art, however, we do think they represent interesting models for our purpose.
We use three models among the top performers in the MTEB.
First, the general text embeddings (GTE) model "gte-large" developed by the Alibaba DAMO Academy~\cite{li2023general}.
This model is trained contrastively in both an unsupervised and a supervised fine-tuning fashion and as of September 2023 leads the MTEB leaderboard for semantic text similarity (STS) tasks.
Second, we use the BAAI General Embeddings (BGE) model "bge-large-en"~\cite{BAAI_2022}.
This model currently tops the overview MTEB leaderboard with minimal background information on the type of model and training available.
Thirdly, we use OpenAI's general purpose embedding model "text-embedding-ada-002" offered at their API since December 2022 \cite{OpenAI_2022, brown2020language}. 
Not many architectural details about this model are known, however, it performs in the top 20 models both in the overview as well as for STS tasks as of September 2023.
Furthermore, it's likely one of the most widely used models in commercial applications.

We extend the model list with two specialized models:
a pure word embedding model for emotion-enriched word embeddings (EWE)~\cite{agrawal2018learning}, trained to mitigate the inductive bias that emotion term embeddings are liable to be influenced by, and Microsoft's cross-modal text-audio embedding model (CLAP)~\cite{elizalde2023clap}, trained to embed both text and audio excerpts in the same space for cross-modal retrieval.
We assume these models to be better suited to the language in the domains of audio and emotion description, respectively, both of which overlap with expressive performance characterization.

The last year has seen several publications investigating retrieval of perceptual structure information from large language models (LLM), and in particular using the model chatGPT.
Most closely related to our proposal are ratings of timbral similarity~\cite{siedenburg2023language}, music similarity~\cite{flexerchatGPT}
, and general sensory judgment dimensions~\cite{marjieh2023large}, all of which find evidence for the recovery of human annotations by chatGPT, albeit not at human level.

%con espressione, music research data
Our reference data is based on free text descriptions of expressive piano performance.
People involved with music take pleasure in talking about music.
This is no different for expressive performance of Western classical solo piano.
In doing so they develop a rich vocabulary that relates to different aspects of performance such as evaluative/axiological terminology, emotions, metaphors, playing technique, or timbre descriptors. 
In music research, these aspects are often addressed separately, and with a generally reductionist approach.
That is, researchers are interested in the identification of underlying factos, perceptual categories, and their relations to acoustic features~\cite{palmer1997music,Lerch-2020}.
This is somewhat opposed to our approach, where no reduction of the semantic space is pursued.
Nevertheless, we want to briefly outline two relevant areas of inquiry: emotions and timbre.

\begin{figure*}[t]
\begin{center}
\includegraphics[trim={0 0 0 0},clip,width=0.95\linewidth]{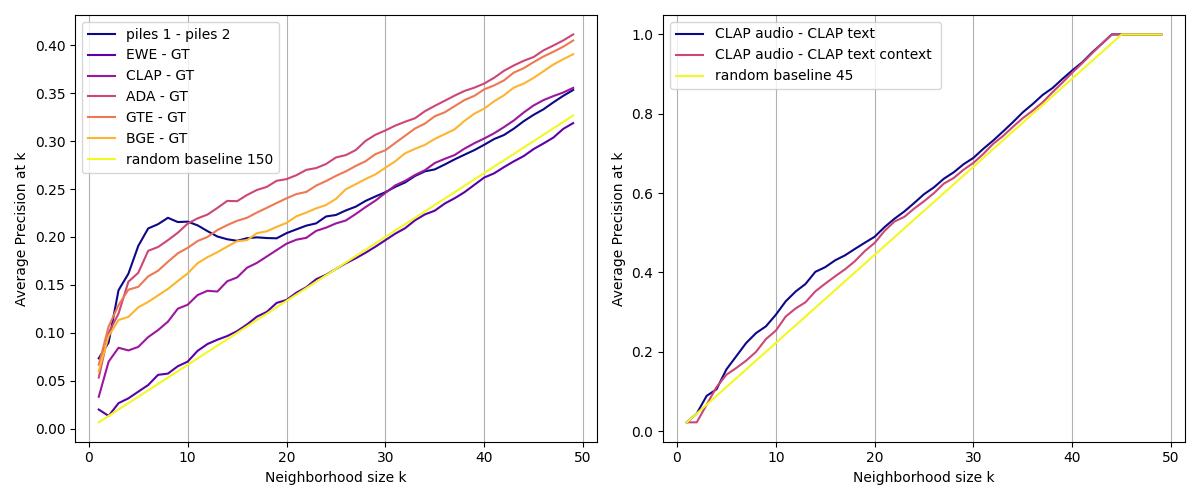}
\caption{Left plot: aP@k for k $\in \{1, ..., 49\}$ for several embeddings models against the ground truth similarities. Right plot:  aP@k of 45 performance embeddings represented as CLAP audio embeddings and as mean CLAP text embeddings of terms (with and without context prompts).}  
\label{fig:prec_at_k}
\end{center}
\end{figure*}

% emotion
There exists a substantial literature investigating the emotional language related to music~\cite{gabrielsson2003music,eerola2012review}.
Among prominent models are categorical models such as the Geneva Emotional Music Scale (GEMS)~\cite{zentner2008emotions} and the dimensional valence and arousal model~\cite{Russell:1980wy}.
Crucial questions relate to the question whether musical emotions are perceived or induced~\cite{song2016perceived} or the paradoxical enjoyment of negative emotions such a sadness~\cite{eerola2018integrative}.
Metaphors are another common linguistic device in the characterization of music.
We refer to Schaerlaken et al.~\cite{schaerlaeken2019hearing} who identified five main factor in their analysis of metaphorical attributed (GEMMES) and connected their perception to the GEMS~\cite{schaerlaeken2022linking}.

%timbre
Timbre is crucial topic in performance research and music psychology~\cite{krumhansl1989musical,saitis2020timbre, esling2018generative}.
Timbre is increasingly conceptualized as controlled expressive performance attribute, i.e., as something that performers can influence with playing techniques and gestures.
Most relevant to our work are several piano timbre description experiments by Bernays et al.~\cite{bernays2013expressive, bernays2014investigating, bernays2010expression,bernays2011verbal} which brought five to eleven categorical terms to the fore.
For our experiments, we rely on as many terms with associated similarity annotations as possible without reduction to principal factors or dimensions, which we find in the con espressione dataset and its pile sorting extension, detailed in Section~\ref{sec:ced}.

\begin{figure*}[t]
\begin{center}
\includegraphics[trim={0 0 0 0},clip,width=0.95\linewidth]{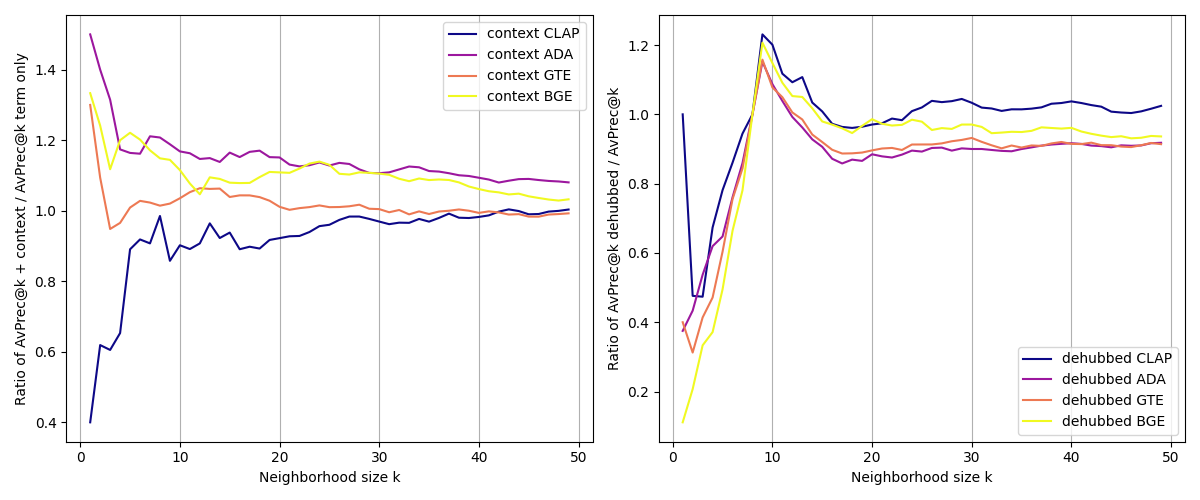}
\caption{Left plot: relative change in aP@k brought about by the inclusion of contextualizing prompts. Right plot: relative change in aP@k due to hubness reduction at neighborhoods of size eight.}  
\label{fig:context_hubness}
\end{center}
\end{figure*}

\section{Methods}

For our experiments we require an embedding space of words (for simplicity: adjectives) with associated ground truth similarity ratings as a type of ground truth reference data.
We also need several SOTA embedding models and the metrics by which we can meaningfully compare the resulting similarity structures.
In this section, we introduce these components.

\subsection{Con Espressione Data}\label{sec:ced}

The Con Espressione Dataset (CED) collected descriptions of piano performances through an online questionnaire~\cite{cancino2020characterization,CED2020}.
Participants listened to 45 different performances by famous pianists of nine excerpts of Western classical piano pieces.
The participants were shown the prompt: 'Please think of words (if possible, adjectives) that best describe the character of each performance to you.' 
and
a text field allowed for free text answers (as many words as they liked, in a language of their choice).
They were further instructed to concentrate on the performance aspects and not on the piece of music itself.
The CED characterizations contains 3,166 terms, of which 1,415 are unique.
Consequently, the CED consists of very loosely structured text data for which relational semantic ground truth, i.e. which answers refer to the same aspect or idea, is largely missing.

To mitigate this, a follow-up experiment was designed~\cite{cancino2021sorting}.
In two separate sessions, groups of professional musicians/musicologists sorted the 150 most often occurring terms in the CED into piles.
These piles should cluster terms that describe a common expressive character.
The type of similarity and number of piles were left open.
The two groups of four professional musicians each sorted the terms into 25 and 19 piles, respectively, and gave a name to each pile, in the form of an adjective that best summarizes the common meaning of the words associated with the pile. 
\footnote{An interactive interface for the exploration of piles, terms, and performances is available at: \url{https://cpjku.github.io/expressivity/}}

\subsection{Embeddings and Similarities}

We use the same 150 terms as our adjectival space.
To derive ground truth similarities between the terms we use both term co-occurrence in piles (from both groups) and co-occurrence for performance descriptions.
For each of the pile groups, we create a similarity matrix where pairwise similarities of terms within a pile are set to one, outside the pile to zero. 
Term similarities are also computed based on the CED directly, where two different terms that occur in the characterization of the same performance are assumed more similar than if they are used for different performance.
We again compute a similarity matrix where co-occurrence of two terms in the same performance description sets the similarity to one, else to zero.
We weigh each source of similarity annotations the same, that is, the similarities from pile group one, pile group two, and performance description are summed up and finally normalized.
Figure~\ref{fig:mds_main} illustrates the resulting similarity structure.
The legend on the left lists all piles of both groups (with the names the musicians gave them) the scatter plot on the right shows a low-dimensional approximation of the term similarities.

We compare our ground truth similarities with similarities for the same adjectival space, i.e., the same 150 terms, embedded using five different embedding models: ADA (embedding dimensionality 1536), CLAP (1024), EWE (300), GTE (1024), and BGE (1024), detailed in Section~\ref{sec:related}.

\subsection{Metrics}

We assess the similarities using several metrics.
All similarities are cosine similarities in the embedding spaces:

\begin{equation}
    CosineSimilarity(x,y) = \frac{x \cdot y}{|x||y|}
\end{equation}

for $x$ and $y$ term embeddings in the same space. 

Our main evaluation metric is the average precision at k (aP@k) for k nearest neighbors.
For each term x in our adjectival space S, we compute two neighborhoods of size k, one according to our ground truth embedding space U (knn(x,U)), and one according to a test embedding space V (knn(x,V)).
We then compute the precision of retrieval of the U-neighborhood by the V-neighborhood:

\begin{equation}
    aP@k(U,V) = \frac{1}{|S|}\sum_{x \in S} \frac{|knn(x,U) \cap knn(x,V)|}{k}
\end{equation}
where knn(x,U) denotes the set of k nearest neighbors of x according to the embedding space U.
We choose aP@k over the more common Spearman correlation of embedding similarities for STS tasks~\cite{muennighoff2022mteb}, for its capacity to naturally differentiate each level of neighborhood size k and its rank-agnostic behavior within the neighborhoods, which corresponds to the ground truth pile sortings.

In Section~\ref{sec:cluster}, we compare the agreements between the pile groups (ground truth sets of clusters) with k-means clusterings in the embedding spaces.
For this purpose, we compute the overlap coefficient between sets of clusters according to different similarities. 
In particular, we compute the average maximal overlap between two different sets of clusters
$\mathbb{C}_1$ and $\mathbb{C}_2$ as:
\begin{equation}
    AvMaxOverlap(C_1,C_2) = \frac{1}{|\mathbb{C}_1|} \sum_{C_1 \in \mathbb{C}_1}  \max_{C_2 \in \mathbb{C}_2}\frac{|C_1 \cap C_2|}{min(|C_1|,|C_2|)}
\end{equation}
Note that this metric is not symmetrical in $\mathbb{C}_1$ and $\mathbb{C}_2$.

\section{Experiments and Results}

This section presents experiments addressing structural correspondences between similarity spaces. 
After a main experiment in Section~\ref{sec:prec}, we investigate more fine-grained issues regarding the correspondence of text and audio embeddings in Section~\ref{sec:audio}
the effects of hubness in the embedding spaces in Section~\ref{sec:hubness}, the effects of contextualized terms in Section~\ref{sec:context}, and the overlap in clusterings in Section~\ref{sec:cluster}.

\subsection{Similarity Structure Correspondence}~\label{sec:prec}

To what extent do the models' embedding spaces correspond to the similarity relations derived from expert annotations?
We answer this question using aP@k values.
However, we first make a few general observations about the distributions of similarities without reference to structural considerations.

Figure~\ref{fig:simdist} shows box plots of the distributions of pairwise similarities in several spaces under scrutiny.
Note that for all language models (labelled as EWE, CLAP, ADA, GTE, and BGE) the embeddings are very similar, with values rarely falling below 0.7 and mean values above 0.8.
This illustrates the inductive bias of the embeddings as these terms are likely to occur in similar and sometimes very particular domains.

Figure~\ref{fig:prec_at_k} (left) shows the performance  of each embedding in terms of its approximation of the neighborhood structure of a reference embedding.
For neighborhood sizes k between one and 49, the aP@k of each embedding space is computed against our ground truth data.
The models are are clearly ranked for all k, from top to bottom: ADA, GTE, BGE, CLAP, and EWE.
A yellow diagonal denotes a random baseline.
With the exception of EWE are all models significantly better than random.
The upper bound of the 95\% confidence interval (not shown) of the random baseline is approximately 0.02 above the baseline itself at k=1 and gets closer to the baseline as k increases.

A blue line indicates the only values that are not compared against the ground truth.
Instead it compares similarities only based on pile group 1 (co-occuring terms in a pile have similarity one, else zero) with similarities only based on pile group 2.
This is added as an indication of where human agreement about similarity might fall, albeit with some caveats:
The values are only really meaningful for k in the approximate size of piles (roughly 5-10 terms/pile).
For smaller k, the piles do not encode greater or lesser similarity of terms within a pile, all values are one and an arbitrary ordering was created by addition of some minimal noise.
For larger k, the piles do not encode greater or lesser similarity for terms beyond the pile, all values are zero and again minimal noise was added for an arbitrary ordering.
It's also not possible to compare these similarities against the ground truth since they were part of the creation of the ground truth which distorts the result unfairly.
The blue line segment for k = 5 - 10 does give an indication that none of the models reach the agreement of two groups of expert annotators.

\subsection{Audio vs.~Text Embeddings}~\label{sec:audio}

The CLAP model is trained to minimize distances between corresponding text and audio embeddings which makes it possible to compare against audio embeddings of the 45 performances that are characterized by the 150 terms in the con espressione data.
The three leftmost distributions in Figure~\ref{fig:simdist} are related to this approach:
"audio - audio" denotes the pairwise similarities between audio embeddings, "s terms - audio" shows cross-modal similarities between 150 individual terms and 45 performances, and "p terms - audio" shows cross-modal similarities between 45 performances averaged from their corresponding term embeddings (the terms used to describe the performance in the CED) and 45 performance recording embeddings.
Figure~\ref{fig:prec_at_k} (right) shows the latter in an aP@k plot; the 45 performances are embedded in both in the text spaces of two different text models (see Section~\ref{sec:context} for a discussion on the "context" model) and in the audio space.
The values are notably lower, for several k the values are not significantly better than the random baseline, indicating that either the audio or the text space do not have the granularity to represent these minute differences.
We conjecture that the audio model is the more likely source of misalignment.
After all, all performance recordings can reasonably be classified as "classical solo piano" which is closer to the level of precision to be expected from an unspecific audio representation model trained on a variety of (non-)musical audio~\cite{elizalde2023clap}.

\subsection{The Effect of Hubness}~\label{sec:hubness}

\begin{table}
\begin{tabular}{lr|lc|lc}
\toprule
\textbf{Model} & nbhd & \textbf{Skewness}  && \textbf{Robinhood} & \\
\toprule
ADA &   4   &    0.99 &    0.88 &    0.25 &    0.21\\%\multirow{5}{*}{4}
CLAP &   4  &    1.09 &    0.52 &    0.22 &    0.19\\
GTE &  4  &    1.11 &    0.91 &    0.24 &    0.20\\
BGE &   4  &    1.76 &    1.14 &    0.32 &    0.29\\
RB &   4  &    0.54 &    0.40 &    0.17 &    0.13\\
\midrule
ADA &    8 &    1.12 &    0.58 &    0.25 &    0.21\\
CLAP &    8 &    1.19 &    0.54 &    0.25 &    0.23\\
GTE &    8 &    1.13 &    0.44 &    0.24 &    0.21\\
BGE &    8 &    1.97 &    1.63 &    0.39 &    0.36\\
RB &    8 &    0.37 &    0.17 &    0.13 &    0.11\\
\midrule
ADA &   16 &    0.81 &    0.32 &    0.26 &    0.24\\
CLAP &   16 &    1.21 &    0.70 &    0.25 &    0.23\\
GTE &   16 &    0.67 &    0.24 &    0.25 &    0.24\\
BGE &   16 &    1.83 &    1.61 &    0.43 &    0.42\\
RB &   16 &    0.25 &    0.04 &    0.10 &    0.08 \\
\bottomrule
\end{tabular}
\caption{Results of hubness measurement and reduction. 
Values for each model (RB random baseline) are presented for three different neighborhood sizes (nbhd). All skewness and robinhood values are doubled, left original, right after hubness reduction.
}\label{tab:hubness}
\end{table}

The existence of hubs has repeatedly been found a source of distorted similarity structures, especially in high-dimensional spaces~\cite{feldbauer2019comprehensive}.
Hubs are points that appear too often in k nearest neighborhoods of other points and as such are liable to influence the aP@k.
In this experiment we address the influence of hubs for four models.

We first measure hubness as both skewness of the k-occurrence histogram (higher skewness indicates more hubness) and as robinhood index (which indicates the percentage of slots in nearest neighbor lists would need to be redistributed for equal distribution).
We carry this computation out for three neighborhood sizes (4,8, and 16) and compute hubness reduction using an approximate mutual proximity method~\cite{schnitzer2012local}.
For algorithmic details regarding hubness measurement and reduction we refer the reader to Feldbauer et al.~\cite{feldbauer2018fast}.

Table~\ref{tab:hubness} shows the results of hubness reduction. 
For each metric (skewness, robinhood) we note two values for each setting: before (left) and after (right) hubness reduction.
Several models show significant hubness (skewness > 1.0, robinhood > 0.25), with BGE being a negative outlier, and they almost universally benefit from reduction.

What does this mean for similarity structure recovery against the ground truth?
Figure~\ref{fig:context_hubness} (right) shows the relative change in aP@k for 4 models after hubness reduction at neighborhood size eight.
This neighborhood was chosen for being crucial against ground truth based on piles, which on average have approximately this size.
Notably, hubness reduction leads to a decrease for k less than the set neighborhood.
For values around eight, hubness reduction universally leads to an increase in performance of about 20\%, enough to boost the highest performing models an the league of the expert agreement baseline in this crucial area (see Figure~\ref{fig:prec_at_k} on the left, see section~\ref{sec:prec}).

\begin{figure}[t]
\includegraphics[trim={0 0 0 0},clip,width=1.05\linewidth]{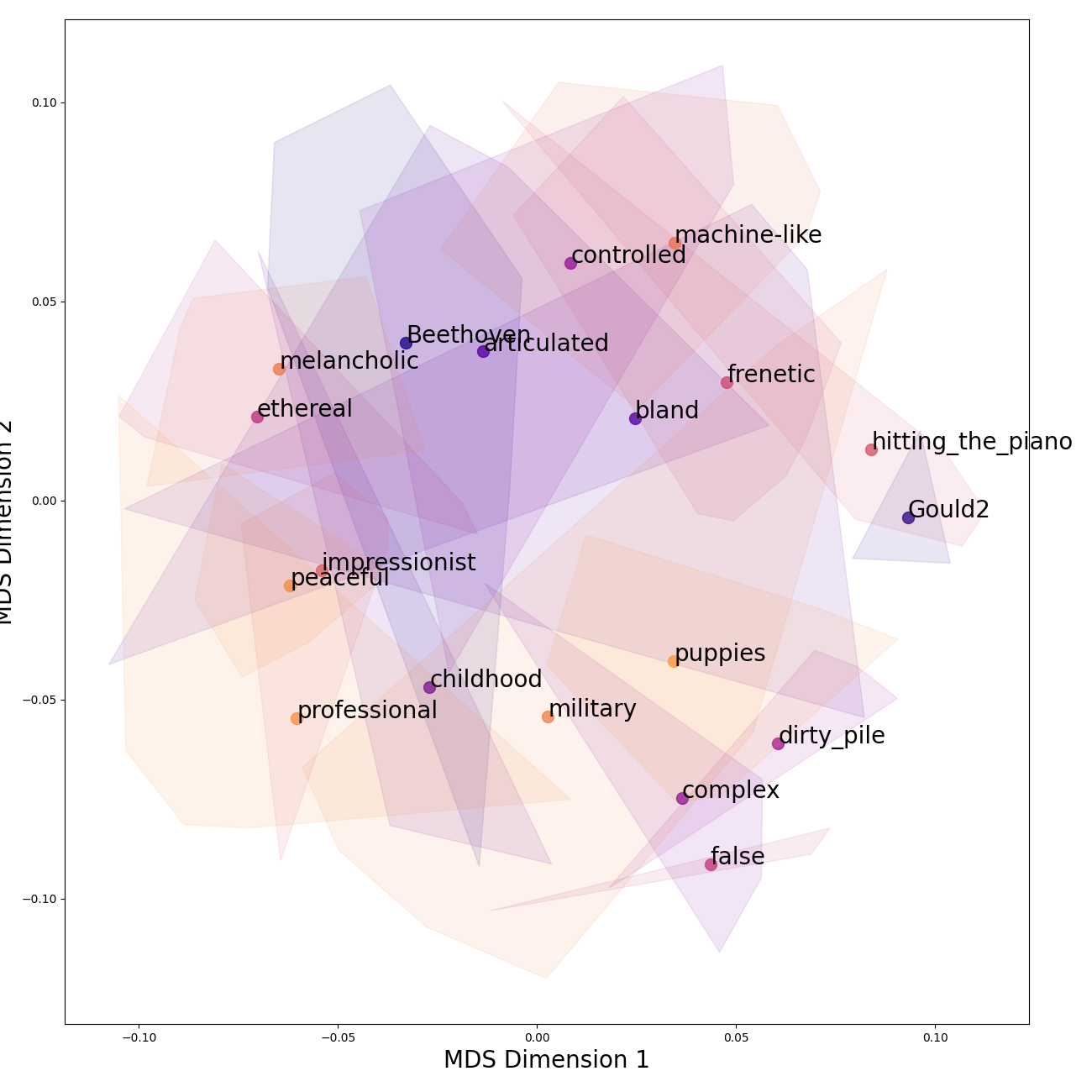}
\caption{Visualization of the convex hull of terms of each pile as embedded in the ground truth data. MDS dimension reduction for illustrative purposes only, 8+ dimensions are required to represent the space with minimal loss of information (<10\% reduction in aP@k against original data).
The pile centers are shown as average term embedding positions and annotated with the pile names given by the musicians.}  
\label{fig:convex_hull}
\end{figure}

\subsection{The Effect of Context}~\label{sec:context}

SOTA text embedding models are often capable of encoding term contexts into their representations, i.e., terms can be augmented with suitable prompts to specify the context of use.
Such contextualizing information ideally enables the model to learn domain-specific similarities and relations that might not be apparent in other situations, such as metaphorical usage of terms related to movement, weight, and flow, or words borrowed from other sensory modalities like sweet, rough, and warm, which are common in our ground truth data.
In this experiment, we augment the 150 terms with a common context prompt.
In the original Con Espressione game, listeners were asked to `please think of words (if possible, adjectives) that best describe the character of each performance to you.'.
We translate this to the prompt: 'I listen to a solo piano performance of a classical piece of music and I'd describe the character of the performance as TERM' and recompute all embeddings for four test models.

Figure~\ref{fig:simdist} shows the distributions of pairwise similarities for context prompts.
For all tested models, the prompts led to higher pairwise similarities.
To test the similarity structure of these context embeddings, we again  compute the relative change in aP@k for four models after adding contexts.
Figure~\ref{fig:context_hubness} (left) shows this relative change, i.e., the aP@k of the embeddings with context divided by the aP@k of those without.
Not all models react positively context information, GTE and CLAP stay largely the same or get worse.
The other two models see performance increases of 20 \% and more, which in the case of ADA pushes the model higher than the expert reference (see Figure~\ref{fig:prec_at_k}, see section~\ref{sec:prec}).

\subsection{Clustering and Piles}\label{sec:cluster}

The aP@k compares neighborhoods of the same size, however, the experts' groups of piles are not homogeneous in size and number.
On the other hand, the piles do provide a complete clustering of the adjectival space which can be compared against automatic clusterings (see Figure~\ref{fig:convex_hull} for an illustration of the clustering provided by pile group two in a the ground truth space).
In this last experiment, we compute k-means clusterings in several embedding spaces and compare them against the two groups of piles by means of average maximal overlap coefficients.

Table~\ref{tab:overlap} shows the results.
All k-means clusterings are computed with k=22, the average of the pile group sizes ($|P1|=25, |P2|=19$).
Average maximal overlap coefficients are not symmetrical, hence we report two values per setup.
The top row reports two values for reference: the average maximal overlap of pile group one with pile group two ("P1 in P2"), and vice versa ("P2 in P1").
K-means clusterings are computed for six models, four embedding models, the ground truth model, and a random baseline consisting of a 100-dimensional Gaussian.
The two columns below "Overlap v P1" average maximal overlap coefficients for group one in k-means clusters (left) and k-means clusters in group one (right).
The next two columns marked "Overlap v P2" report the same values for group two.

Note that smaller sets of clusters generally reach a higher average maximal overlap due to larger clusters (P2 in P1 > P1 in P2, $|P1| = 25$, $|P2| = 19$).
None of the embedding models reach the agreement between the two groups of piles, however, they clearly outperform the random baseline.
The ground truth reaches higher overlap values with both groups than in between the groups, which is to be expected, as it was derived from the performance annotations and the two groups of piles.

\begin{table}
\begin{tabular}{l|lc|lc}
\toprule
 Ref & P1 in P2: & 0.62     &P2 in P1:&0.65   \\
   \midrule
   \textbf{Model} & \multicolumn{2}{l}{\textbf{Overlap v P1}} &  
   \multicolumn{2}{l}{\textbf{Overlap v P2}}  
    \\
   \midrule
   ADA & 0.55 &    0.58 &    0.48 &    0.43 \\
   CLAP & 0.52 &    0.51 &    0.47 &    0.46 \\
   GTE & 0.59 &    0.59 &    0.53 &    0.52 \\
  BGE &  0.51 &    0.53 &    0.53 &    0.48 \\
  GT &  0.66 &    0.70 &    0.66 &    0.70 \\
  RB &  0.39 &    0.41 &    0.40 &    0.39 \\
\bottomrule
\end{tabular}
\caption{Overlap and distance ratios. 
}\label{tab:overlap}
\end{table}

\section{Discussion and Conclusion}

% start from data again
Specialized datasets created by domain experts like the CED and its sorted pile groups usually serve a primary reductionist research purpose: the identification and description of dimensions, categories, and relations in perceptual-linguistic spaces.
However, they also allow for a quantitative glimpse into the similarity structures of these spaces:
similarity structures which are hypothesized to be recovered by SOTA term embedding models.
In a series of experiments, we address this hypothesis for a presumably highly specialized type of adjectival space, that of characterizations of expressive performance of Western classical solo piano works.

% summarize results
Our results show that domain-specific semantic similarity structures are indeed represented in the embedding spaces --- to a degree.
The tested models span the full range from near the random baseline to near human agreement.  
General-purpose models perform better than domain-adapted ones, running counter to our initial assumption.
% ADD one more sentence on this

In our experiments, cross-modal audio embeddings of the performance recordings fail to exhibit the same similarity structure as text embeddings.
Hubness reduction helps the similarity correspondence universally, albeit only for a specific and small segment of neighborhoods.
The inclusion of contextualizing prompts affects the models differently, with the best models receiving a clear boost in quality.
Overlap statistics show that all embedding space clusters show less correspondence with the expert sortings than those show among themselves.

% reliability of data
The groups of piles are used as a reference for correspondence in similarity structures throughout as they represent a rough estimate of inter-rater agreement to be expected.
They are however the largest source of uncertainty.
We do not know how likely similar groups of piles are or have other means of assessing of inter-rater agreement.
Neither do we have direct similarity ratings or know whether performance expressivity-specific similarities are indeed notably different from general similarity.  
Research into dimensional and categorical structure on music perception as discussed in section~\ref{sec:related} may ground the pile group similarities, however, for the number of terms used or even the free-text CED annotations more research is required to illuminate the robustness of this data.

% conclude
To conclude, general state-of-the-art text embedding models can show correspondence with expert annotated perceptual-linguistic similarities that reach the experts' inter-rater agreement while other --- even plausibly better suited domain models --- fail at this task.
Future research includes the investigation of the robustness of the annotation data as well as the extension of this approach to other domains where fine-grained and possibly idiosyncratic adjectival spaces are used.

\section{Reproducibility}

Our data and code is available at: 

\noindent\url{https://github.com/CPJKU/performance_embeddings_fire23}

\begin{acks}
This work is supported by the European Research Council (ERC) under the EU’s Horizon 2020 research \& innovation programme, grant agreement No.~101019375 (“Whither Music?”), and the Federal State of Upper Austria (LIT AI Lab).
\end{acks}

\bibliographystyle{ACM-Reference-Format}
\bibliography{sdp-bib-base}
\end{document}